\documentclass[acmlarge]{acmart}
\AtBeginDocument{%
  \providecommand\BibTeX{{%
    \normalfont B\kern-0.5em{\scshape i\kern-0.25em b}\kern-0.8em\TeX}}}

\acmConference[BCB]{ACM Conference on Bioinformatics, Computational Biology, and Health Informatics}{August 07--10, 2022}{Chicago, IL}
\begin{document}

\title{Identifying Dementia Subtypes with Electronic Health Records}

\author{Sayantan Kumar}
\email{sayantan.kumar@wustl.edu}
\orcid{}

\affiliation{%
  \institution{Department of Computer Science and Engineering, Washington University in St. Louis}
  \city{St. Louis}
  \state{Missouri}
  \country{USA}
}

\author{Zachary Abrams}
\affiliation{%
  \institution{Institute for Informatics, Washington University School of Medicine}
  \city{St. Louis}
  \state{Missouri}
  \country{USA}
}

\author{Suzanne E. Schindler}
\affiliation{%
 \institution{Department of Neurology, Washington University School of Medicine}
 \city{St. Louis}
 \state{Missouri}
 \country{USA}}
 
\author{Nupur Ghoshal}
\affiliation{%
 \institution{Department of Neurology, Washington University School of Medicine}
 \city{St. Louis}
 \state{Missouri}
 \country{USA}}

\author{Philip R. O. Payne}
\affiliation{%
  \institution{Institute for Informatics, Washington University School of Medicine}
  \city{St. Louis}
  \state{Missouri}
  \country{USA}
}




\renewcommand{\shortauthors}{Kumar et al.}

\begin{abstract}

Dementia is characterized by a decline in memory and thinking that is significant enough to impair function in activities of daily living. Patients seen in dementia specialty clinics are highly heterogeneous with a variety of different symptoms that progress at different rates. In this work, we used an unsupervised data-driven K-Means clustering approach on the component scores of the Clinical Dementia Rating (CDR®) score to identify dementia subtypes and used the gap-statistic to identify the optimal number of clusters. Our goal was to characterize the identified dementia subtypes in terms of their cognitive performance and analyze how patient transitions between subtypes relate to disease progression. Our results indicate both inter-subtype variability, which indicates the variability amongst dementia subtypes for a particular component score even with the same CDR and (ii) intra-subtype variability, which indicates the variation in the 6 component scores within a particular dementia subtype. We observed that dementia subtypes that represented individuals with very mild dementia (CDR 0.5) had widely varying rates of transition to other subtypes. Future work includes testing the generalizability of our proposed pipeline on additional datasets, and using a larger volume of EHR data to estimate probabilistic estimates of the variability between dementia subtypes both in terms of cognitive profile and disease progression.

\end{abstract}

\keywords{Dementia, electronic health records, heterogeneity, clustering, patient subtypes, interpretability, disease progression}


\maketitle

\begin{figure*}[htbp]
    \includegraphics[width = \linewidth]{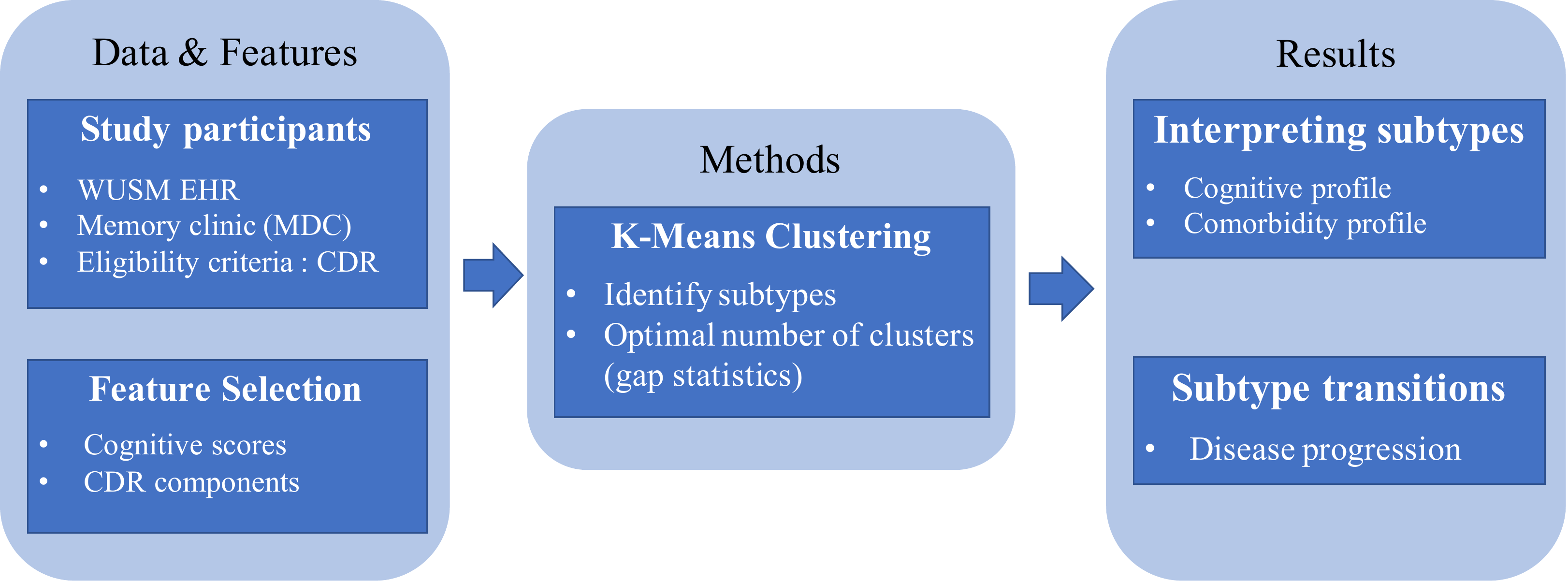}
    \caption{High-level workflow of our proposed approach. First, we identified outpatient office visits from the Electronic Health Records data and included patients whose visits record a Clinical Dementia Rating (CDR) score. For feature variables, we extracted the cognitive assessment scores and the CDR component scores of patients. The K-Means clustering algorithm was applied to find the subtypes and the optimal number of clusters was determined by the gap-statistics algorithm. Next, we characterize the identified dementia subtypes in terms of their cognitive performance and analyze how patient transitions between subtypes relate to disease progression. Keys: WUSM - Washington University School of Medicine, EHR - Electronic Health Records, MDC - Memory Diagnostic Center, CDR - Clinical Dementia Rating.}
    \label{fig:workflow}
\end{figure*}

\section{Introduction}
\label{sec:intro}

Dementia is defined by a decline in memory and thinking that is significant enough to impair function in activities of daily living. Dementia has numerous causes, including reversible causes, such as medication-induced cognitive dysfunction, as well as irreversible causes, such as progressive neuro-degenerative disorders \cite{ferri2005global}. Alzheimer Disease (AD) is the most common cause of dementia in older adults, but many other disorders can cause or contribute to dementia, including cerebrovascular disease and diseases associated with Lewy bodies, tau tangles, or TARDNA-binding protein43 (TDP-43) \cite{zanetti2006mild}. Patients present with a variety of different symptoms and progress at different rates, which may be related to the underlying brain pathologies, variation in baseline cognitive ability, genetic background, medical comorbidities, and social determinants of health \cite{ryan2018phenotypic}. Hence, patients seen in a dementia clinic are highly heterogeneous and represent many different types of dementia. Better characterizing the clinical heterogeneity of dementia could improve dementia diagnosis and improve the ability of clinicians to provide an appropriate prognosis to patients and their families \cite{goyal2018characterizing}.  

Previous studies have used a priori-defined neuropathological categories to identify subtypes within a clinical dementia population \cite{jack2015different, lam2013clinical, murray2011neuropathologically}. However, these approaches define subtypes of pathologies based on clinical diagnosis, which rely on a "clinical intuition" instead of unbiased data-driven approaches \cite{nettiksimmons2014biological, noh2014anatomical}. More recently, the increasing availability of Electronic Health Records (EHR) data combined with complex machine learning algorithms have encouraged data-driven approaches to identify patterns in clinical data \cite{kumar2021machine}. These new data sources and methods may offer new insights into the underlying heterogeneity of dementia. For example, clustering algorithms can stratify dementia patients into subtypes based on key features recorded in the EHR that can enhance predictive ability compared to analyzing the entire cohort as a single homogeneous group \cite{vogt1992cluster}. 

There has been significant research which uses clustering techniques on EHR data to identify dementia subtypes. In \cite{xu2020data}, hierarchical clustering was applied on clinical data from a multi-speciality urban medical centre to identify multiple sub-phenotypes of Alzheimer Disease and related dementia. Another study applied a representation learning model on Mount Sinai Health system data and identified  subtypes with variable degree of dementia symptoms \cite{landi2020deep}. Other studies focusing on dementia subtyping used 2 categories of structured clinical data sources as key features : brain atrophy patterns as measured by structural MRI \cite{varol2017hydra,dong2015chimera, poulakis2018heterogeneous,forstl1994pathways,malpas2016structural} and cognitive impairment as measured by performance on cognitive tasks \cite{scheltens2017cognitive, scheltens2016identification,wallin2011galantamine,price2015dissociating,davidson2010exploration}. While most of these studies focus on identifying subtypes within a dementia cohort, little research has been done on interpreting the cognitive characteristics of the identified subtypes and how they are related to disease progression.

In this work, we use an unsupervised data-driven clustering approach to identify dementia subtypes. Our goal is to analyze if the identified subtypes have a logical relationship to each other based on our clinical understanding and knowledge. In contrary to prior work, our work focuses on both the interpretation of the identified subtypes based on domain knowledge and analyzing how the subtypes play a role in the longitudinal progression trajectory of the disease. Understanding the cognitive profile of the subtypes can lead to effective clinical decision-making and precision diagnostics tailored to each subtype. Our contributions can be summarized as follows (visual representation of the pipeline shown in Figure \ref{fig:workflow}): 

\begin{itemize}
    \item Applying unsupervised clustering techniques on cognitive assessment scores to identify subtypes within a clinical dementia cohort.
    \item Interpretation of the identified subtypes in terms of their cognitive characteristics.
    \item analyze transitions between the different subtypes and how they are related to disease progression.
\end{itemize}


\begin{table}[htbp]
\caption{Demographic characteristics of our selected cohort (N = 1845 patients). All feature variables were calculated at the baseline visit of each patient.}
\label{tab:demo_table}
\begin{tabular}{@{}ll@{}}
\toprule
Feature variables (baseline) & \\ \midrule
Age (mean, SD)               & 73.2 years (11.9 years)              \\ 
Female (n, \%)               & 1038 (57\%)                          \\ \\
\begin{tabular}[c]{@{}l@{}}Race (n, \%)\\       -White\\       -Black or African American\\       -Asian\end{tabular} &
  \begin{tabular}[c]{@{}l@{}}\\ 1605 (88.9\%)\\ 181 (10\%)\\ 20 (1.2\%)\end{tabular} \\ \\
\begin{tabular}[c]{@{}l@{}}Clinical Dementia Rating (Median, IQR)\\       -Memory   \\       -Orientation\\       -Judgement   and Problem Solving\\       -Community   Affairs\\       -Home   and Hobbies\\       -Personal   Care\end{tabular} &
  \begin{tabular}[c]{@{}l@{}}\\ 1 (0.5, 1)\\ 0.5 (0, 1)\\ 0.5 (0, 1)\\ 0.5 (0, 1)\\ 0.5 (0, 1)\\ 0 (0, 1)\end{tabular} \\ \\
\begin{tabular}[c]{@{}l@{}}Cognitive Characteristics (Median, IQR)\\       -Mini-Mental   State Exam  \\       -Boston   Naming Test\\       -Short   Blessed\\       -Verbal   Fluency\\       -Word   List Recall\\       -Word   List Memory Task\end{tabular} &
  \begin{tabular}[c]{@{}l@{}}\\ 23 (19, 27)\\ 14 (12, 15)\\ 10 (2, 16)\\ 12 (9, 16)\\ 2 (1, 5)\\ 13 (10, 16)\end{tabular} \\ \\
\begin{tabular}[c]{@{}l@{}}Most Common Diagnoses (n, \%)\\       -Memory   Loss\\       -Alzheimer   Disease\\       -Parkinsonism\\       -Major Depressive Disorder\\       -Obstructive Sleep Apnea\end{tabular} &
  \begin{tabular}[c]{@{}l@{}}\\ 1010 (19.6\%)\\ 1075 (9.8\%)\\ 66 (0.7\%)\\ 466 (2.9\%)\\ 374 (2.6\%)\end{tabular} \\ \bottomrule
\end{tabular}
\end{table}


\section{Methods}

\subsection{Data sources and study participants}

Clinical data corresponding to office visits were extracted from the Electronic Health Records (EHR) of patients treated between June 2012 and May 2018 at the Memory Diagnostic Centre (MDC) at the Washington University School of Medicine in St. Louis, a large, academic, tertiary-care referral centre. All patients seen in the clinic were presented for evaluation of memory and/or thinking concerns. Patients underwent a comprehensive history, neurological examination, brief cognitive testing, and if indicated, laboratory testing and brain imaging. Patients were evaluated using the well-validated and widely used Clinical Dementia Rating (CDR®) at each visit, a five-point scale to characterize six different domains of cognitive and functional performance: memory, orientation, judgement and problem solving, home and hobbies, community affairs, and personal care \cite{morris1991clinical}. For both the CDR global score and component scores, 0 denotes cognitively normal, 0.5 very mild dementia, 1 mild dementia, 2 moderate dementia and 3 severe dementia. Longitudinal data from 1,845 patients with 2,737 visits were eligible for inclusion, where each visit had a CDR score. Table 1 represents the baseline demographic characteristics of the selected cohort. This study was approved by the Washington University in St. Louis Institutional Review Board.

\begin{figure*}[htbp]
    \includegraphics[width = \linewidth]{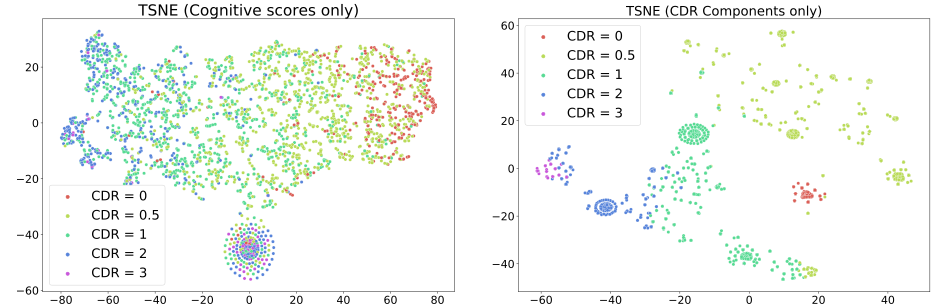}
    \caption{The 2D T-SNE representations of the data for each of the 2 feature categories: cognitive scores only as features (left) and CDR components only as features (right). Each points represents a visit with the colour indicating the Global CDR score.}
    \label{fig:tsne_all}
\end{figure*}

\subsection{Feature selection for clustering analysis}

Our initial set of feature variables consisted of cognitive assessment scores, indicating performance on a standardized battery of cognitive tasks to assess the severity of dementia. Compared to expensive and/or invasive procedures like neuroimaging biomarkers, these scores are standard metrics in dementia research and is recorded for all patients in the memory clinic. Next, we performed feature selection to determine the optimal set of features for clustering. We started with two sets of features as follows: (i) Cognitive assessment scores: Boston Naming Test, Mini-Mental State Exam, Short Blessed, Verbal Fluency, Word List Memory Task and Word List Recall, (ii) six components of CDR score: Memory, Orientation, Judgment and Problem Solving, Community Affairs, Home and Hobbies, and Personal Care. The motivation behind choosing the individual CDR component as features compared to the aggregate Global CDR score is the fact that for patients having the same CDR score, the individual components might be different from one another, allowing us a more granular approach of studying the sub-phenotyping of patients.

For each of the categories, we plotted the t-distributed stochastic neighbour embedding (T-SNE) distribution. Figure \ref{fig:tsne_all} shows the T-SNE 2-dimensional representations of each of the 2 feature sets across the CDR categories. We can see that using the cognitive scores as features creates a gradient from low to high CDR but did not result in clusters. On the other hand, using the CDR components as features created clusters that were distinct without any significant overlap across the CDR categories. analyzing the degree of overlap of points across the different CDR categories, the six components of CDR were selected as the optimal set of features for clustering. Since each of the selected features are ordinal variables and had a very low missing rate of <5\%,  we followed the strategy in \cite{zhou2021imputehr} and imputed the missing values for each feature column using the median value of that feature across all the 2737 visits.

\subsection{Clustering method}

In this study, we used the K-means clustering algorithm to generate the dementia subtypes. In K-means, the a \textit{priori}-specified k number of clusters are identified through iteratively minimizing the distance between data points and their assigned cluster means \cite{likas2003global}. We used the t-distributed stochastic neighbour embedding (T-SNE) \cite{van2008visualizing} as a dimensionality reduction method prior to clustering. T-SNE lowers the dimension of the data while representing points in a geometrical space, thus transforming the data into a form where K-means can be applied. 

One important issue is that the clustering analysis was performed on all data that was available from a 6-year period between 2012-2018, and some patients had multiple visits during this period. For the clustering step, each visit was assumed independent without any temporality (linkage) between individual visits of the same patient. The temporality information between individual visits was used downstream for analyzing the relationship between different clusters. This approach – including all visits for the clustering analysis as opposed to including the baseline visit of each patient was taken to enable downstream longitudinal analysis and track the symptom progression rate among patients. For example, at any given point in time, a patient exists in a single cluster (dementia subtype), but transitions between different clusters over time. The gap statistics algorithm \cite{tibshirani2001estimating} was used for determining the optimal number of clusters (K).  

To gain an insight into which patients have a higher probability of progression, patient transitions between different dementia subtypes were analyzed across multiple visits. Patients with only a single visit were censored from our analysis. All the identified subtypes were characterized in terms of their cognitive characteristics. Finally, the differences in progression rate, both within and between Global CDR score categories were measured.

\begin{figure*}[htbp]
    \includegraphics[width = \linewidth]{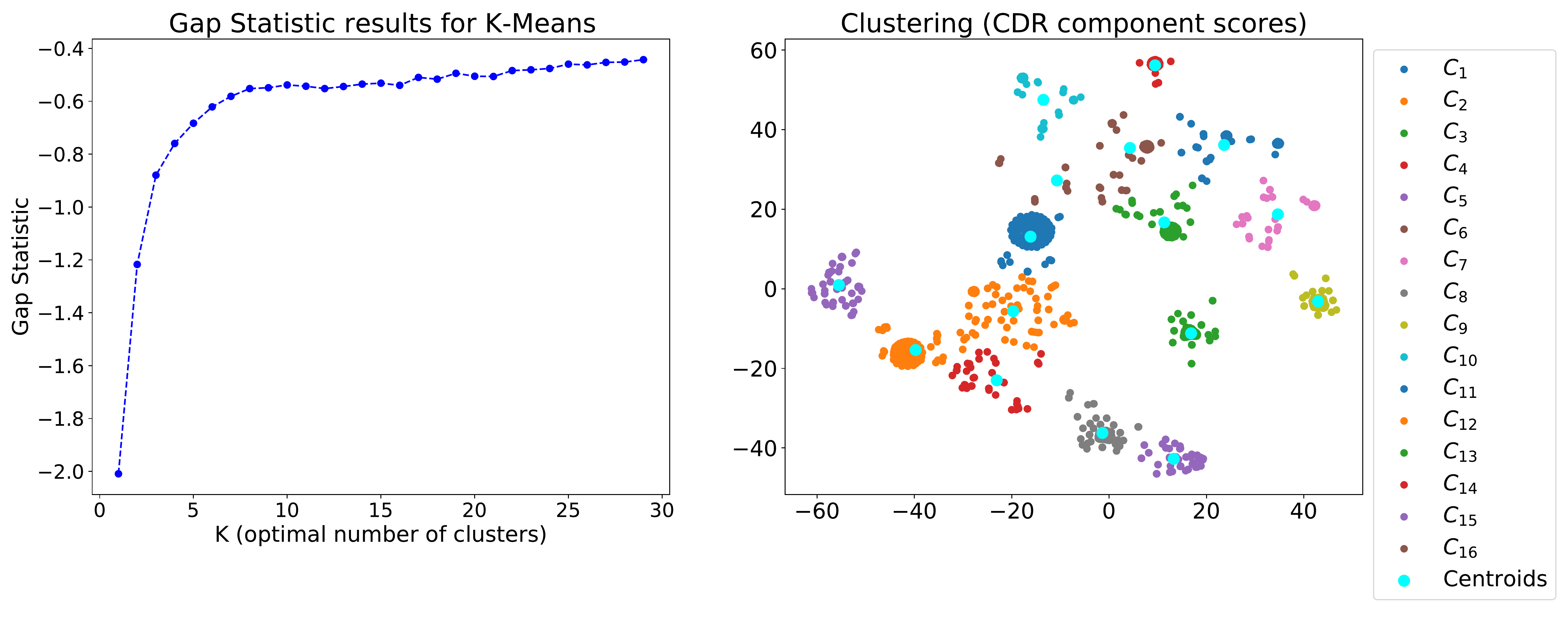}
    \caption{The gap statistics results (left) and the clustering results using CDR component score as features (right). From the gap statistic plot, K = 9, 11 and 16 were identified with significant relative gap jump compared to the other values of K. We selected K=16 as the optimal number of clusters. The points in cyan (right) represent the centroids of each of the identified clusters. Each point represents an individual visit.}
    \label{fig:gap_stat_clustering}
\end{figure*}


\section{Results}

\subsection{Clustering results}

Figure \ref{fig:gap_stat_clustering} shows the gap statistics (left) and the clustering results (right). The gap statistic is used for determining the number of significant components, which was determined by the largest relative gap jump. From the gap statistic plot, K = 9, 11 and 16 were identified with significant relative gap jump compared to the other values of K (Figure \ref{fig:gap_stat_clustering}). In line with the T-SNE distribution of CDR component scores across the CDR categories in Figure \ref{fig:tsne_all}, the final K value was selected as 16. Comparing both the T-SNE and the clustering plots in Figure \ref{fig:tsne_all} (left) and Figure \ref{fig:gap_stat_clustering} (right) respectively, we can observe that the points with CDR = 0.5 are quite scattered and multiple clusters have been formed from them. This validates our hypothesis that dementia patients are quite heterogeneous, even within a single Global CDR level.

\begin{figure*}[htbp]
    \includegraphics[width = \linewidth]{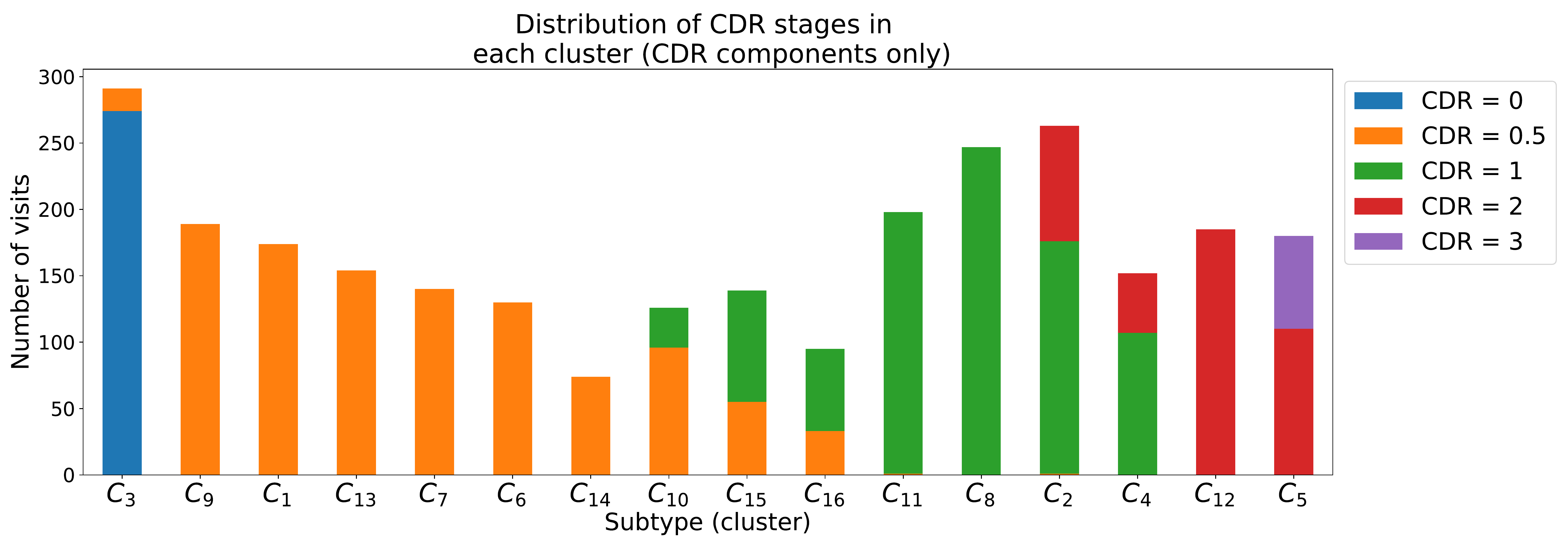}
    \caption{Stacked bar plot showing the CDR composition of each dementia subtypes ordered by increasing CDR score (dementia severity increases moving from left to right). The x-axis shows the 16 dementia subtypes and the y-axis represents the number of visits within each dementia subtypes. Some dementia subtypes included a unique Global CDR  (e.g. $C_1$, $C_{13}$), while other dementia subtypes included two Global CDR scores (e.g. $C_{10}$, $C_{15}$).}
    \label{fig:stacked_cdr}
\end{figure*}

Figure \ref{fig:stacked_cdr} shows the CDR composition of each dementia subtype ordered by increasing CDR score (more severe dementia). Subtypes can either be homogenous, having a unique Global CDR or composite, including two Global CDR scores. We can observe that there is greater variability in early dementia (CDR of 0.5 or 1) leading to more dementia subtypes with a lower CDR score, compared to the more advanced dementia (CDR = 2 and 3). The above finding can also be confirmed from the T-SNE plot in Figure \ref{fig:tsne_all}, where data points with CDR = 0.5 and 1 are more scattered, potentially forming multiple clusters (dementia subtypes). 

\subsection{Cognitive profile of identified subgroups}

The association between the six CDR components and the dementia subtype allows interpretation of the cognitive profile of the subtypes. Figure 5 shows how each of the six components of CDR score vary across the 16 dementia subtypes. The individual component scores vary from 0-3 (0, 0.5, 1, 2, 3), with 0 indicating normal cognition and 3 indicating maximum impairment). For all the 6 CDR components, there was a natural transition of increasing cognitive impairment from the early (mild) to later (moderate to severe) stages of dementia. In this section, we analyzed 2 types of variability: (i) inter-subtype variability, which indicates the variability amongst dementia subtypes for a particular component score even with the same CDR and (ii) intra-subtype variability which indicates the variation in the 6 component scores within a particular dementia subtype.

\subsubsection{Inter-subgroup variability}

\begin{figure*}[htbp]
    \includegraphics[width = \linewidth]{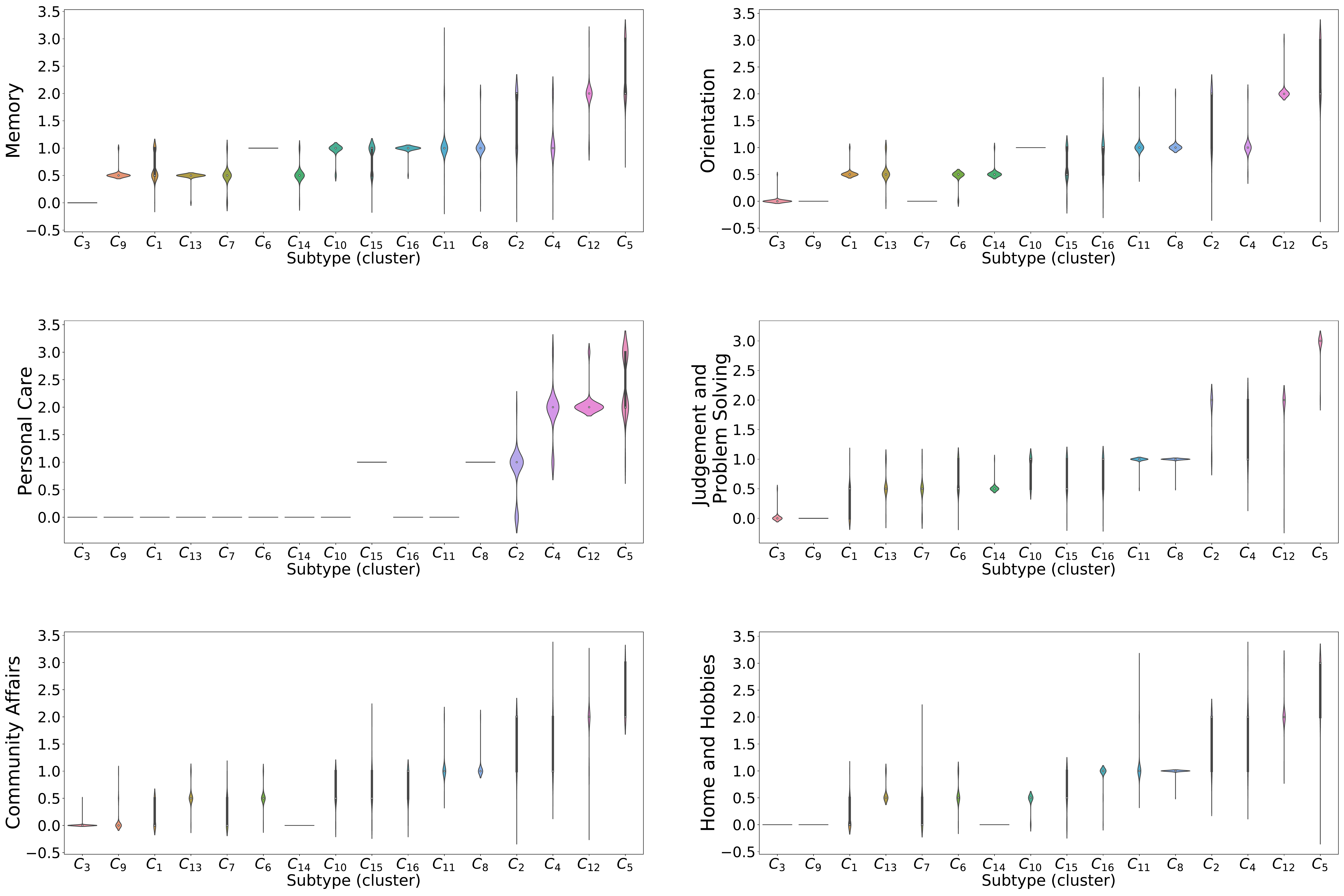}
    \caption{Violin plots showing how each of the how each of the 6 components of CDR score vary across the 16 dementia subtypes. The x-axis represents the individual dementia subtypes and the y-axis shows the component scores varying from 0-3, with 0 indicating normal cognition and 3 indicating severe impairment.}
    \label{fig:cdr_components_cluster}
\end{figure*}

We evaluated the 6 dementia subtypes that only included patients rated CDR 0.5 ($C_1$, $C_6$, $C_7$, $C_9$, $C_{13}$, $C_{14}$), based on our observation in Figure \ref{fig:tsne_all} and \ref{fig:gap_stat_clustering} that the visits with CDR = 0.5 have the maximum underlying heterogeneity. C6 has a consistent memory score of 1 and has more impaired memory loss compared to the remaining 5 dementia subtypes with a mean memory score of 0.5. Similarly, $C_7$ and $C_9$ have fully healthy orientation while all the visits in $C_{10}$ have moderate difficulty with their orientation. Compared to the remaining 5 dementia subtypes, $C_9$ have relatively healthy judgement in solving daily activities. Patient visits in $C_9$ and $C_{14}$ are fully capable of community affairs with well-maintained home and hobbies, compared to the remaining dementia subtypes with relatively impaired performance in these 2 CDR component categories. Unlike the remaining 5 CDR components, personal care had no zero inter-subtype variability with all visits with CDR = 0.5 being fully capable of self-care without any kind of impairment. Dementia subtypes with greater than or equal to mild dementia stages (CDR = 1, 2, 3) had less cognitive variability than subtypes with CDR = 0.5 as observed from Figure \ref{fig:cdr_components_cluster}.

\subsubsection{Intra-subgroup variability}

Next we examined how visits in the same dementia subtype can have variable CDR component distribution with respect to the 6 components of CDR. For example, patient visits in $C_9$ have mild memory impairment, but are otherwise fully functional in the other domains. Patient visits in $C_{14}$ have slightly impaired memory, orientation and problem-solving skills, but are otherwise healthy in personal care, community affairs and home and hobbies. Similarly, we can define the cognitive profile of each dementia subtype, which in turn captures the role of each of the 6 CDR components in capturing the heterogeneity within dementia patients. Analyzing the intra-subtype variability of all the dementia subtypes, we can conclude that memory, orientation and home and hobbies have a significant role in capturing the variability within the CDR = 0.5 groups. On the other hand, the personal care scores do not provide any significant insights.

\begin{figure*}[htbp]
    \includegraphics[width = \linewidth, height = 6 cm]{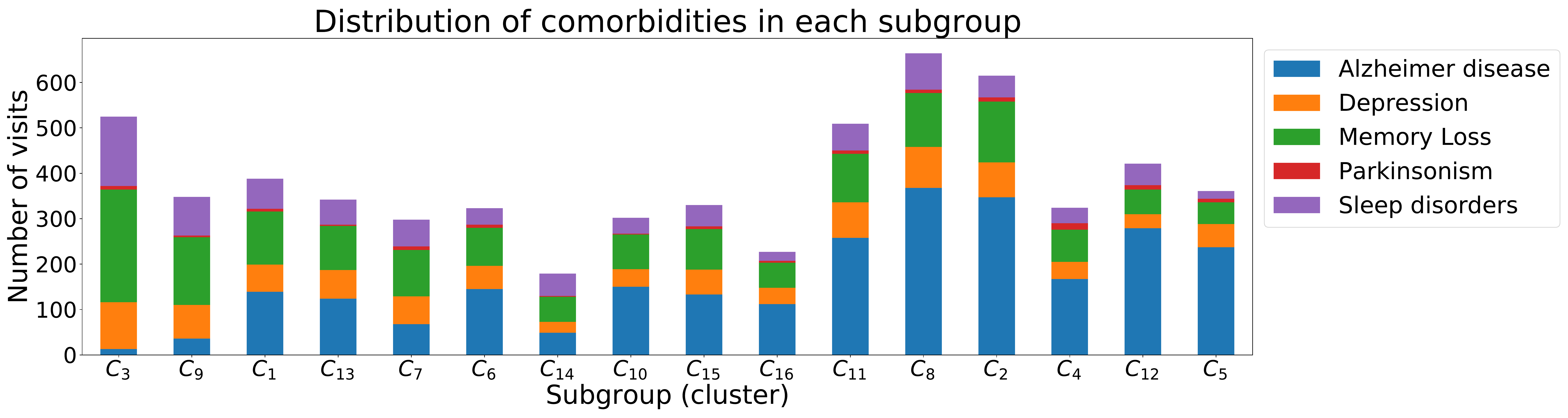}
    \caption{Stacked bar plot showing the comorbidity characteristics of each of the 16 dementia subtypes. The x-axis shows the 16 dementia subtypes and the y-axis represents the number of visits within each subtype having each of the 5 comorbidities.}
    \label{fig:como}
\end{figure*}

\subsection{Comorbidity characteristics of identified subgroups}

We also analyzed the comorbidity profile of each of the 16 identified dementia subtypes. We identified the top 5 comorbid conditions of dementia patients in our dataset, which were Alzheimer Disease (AD), Major Depressive Disorder (depression), Memory Loss (not otherwise specified), Parkinson Disease and Obstructive Sleep Apnea (sleep disorder). Figure 6 shows the distribution of visits within each dementia subtype having the above 5 comorbid conditions. One important trend that we can observe is that the proportion of visits having AD is higher in the later stages of dementia (CDR = 1, 2, 3) compared to the early stage groups (CDR = 0.5). This observation shows that AD, the most common form of dementia, is more likely to be diagnosed in patients with more significant cognitive and functional impairment. On the other hand, the visits in early stage dementia subtypes (CDR = 0.5) are mostly characterized by a diagnosis of memory loss, depression and sleep disorders.

\subsection{Disease progression through subgroup transitions}

We next examined how patients transitioned between the different dementia subtypes over time and how these transitions were related to dementia progression. The units of each subtype are patient visits and a patient can exist in a single subtype at any given point in time, but can transition to the same or different subtype in their next visit. Since there are 16 subtypes and it is difficult to interpret the transitions between every pair of subtypes, we grouped the 16 subtypes into 6 broader categories $G_0-G_5$ ordered as shown in Table 2. The subtypes were grouped based on if they represent a unique Global CDR score (homogenous) or two Global CDR (composite) and also in terms of increasing Global CDR from 0 to 3. For example, $G_0$ represents mostly the CDR = 0 visits while $G_1$ has all the subtypes with CDR = 0.5, corresponding the homogenous orange bars in Figure \ref{fig:stacked_cdr}. 

\begin{table}[h!]
\caption{Grouping the 16 subtypes into 6 broad subtype categories $G_0-G_5$. The subtypes were grouped based on if they represent a unique Global CDR score (homogenous) or two Global CDR (composite) and also in terms of increasing Global CDR from 0 to 3.}
\label{tab:subtype_group}
\begin{tabular}{@{}llll@{}}
\toprule
Grouping & Subgroups                  & CDR     & CDR Type   \\ \midrule
$G_0$       & $C_3$                         & 0 / 0.5 & Composite  \\
$G_1$       & $C_1$, $C_6$, $C_7$, $C_9$, $C_{13}$, $C_{14}$ & 0.5     & Homogenous \\
$G_2$       & $C_{10}$, $C_{15}$,   $C_{16}$            & 0.5 / 1 & Composite  \\
$G_3$       & $C_8$, $C_{11}$                    & 1       & Homogenous \\
$G_4$       & $C_2$, $C_4$                     & 1 / 2   & Composite  \\
$G_5$       & $C_5$, $C_{12}$                    & 2 /3    & Composite  \\ \bottomrule
\end{tabular}
\end{table}

Table \ref{tab:group_transition} shows the number of transitions between every pair of dementia subtype groups. For example, the 170 transitions between $G_1 \rightarrow G_1$ signifies that there are 170 instances where a particular patient that was in $G_1$ (CDR = 0.5) remained in $G_1$ (CDR = 0.5) at their next visit. The transitions involving definite change in CDR include both progression to higher CDR score (e.g. $G_1 \rightarrow G_3$) and regression to a lower CDR score (e.g. $G_4 \rightarrow G_1$). The regressions signify that there are some dementia patients whose cognitive conditions improved in their next visit. While regression is not common, it is important to identify these patients and investigate the reasons for their cognitive improvement. From Table \ref{tab:group_transition}, we can observe that 2 patients progressed from $G_0 \rightarrow G_5$ and $G_1 \rightarrow G_5$ respectively, moving from CDR = 0.5 to CDR = 2, 3 at their next visit and skipping the intermediate CDR = 1 stage, suggesting either rapid progression or a long follow-up interval between visits. On further introspection, we found that the patient moving from $G_0 \rightarrow G_5$ had a short follow-up length (interval between successive visits) of 184 days while the patient moving from $G_1 \rightarrow G_5$ had a long follow-up length of 679 days. While the former observation indicates chances of actual rapid progression, the latter signifies that the probable reason of CDR jump is long follow-up of that patient.

\begin{table}[]
\caption{Number of transitions between every pair of dementia subtype groups. The subtype groups in header column (Start) and header row (End) represent source and target respectively. }
\label{tab:group_transition}
\begin{tabular}{@{}lllllll@{}}
\toprule
Start/End & $G_0$ & $G_1$  & $G_2$ & $G_3$ & $G_4$ & $G_5$ \\ \midrule
$G_0$        & 36 & 15  & 1  & 0  & 0  & 1  \\
$G_1$        & 17 & 170 & 47 & 55 & 13 & 1  \\
$G_2$        & 1  & 19  & 44 & 66 & 24 & 10 \\
$G_3$        & 0  & 5   & 10 & 74 & 51 & 20 \\
$G_4$        & 0  & 2   & 4  & 9  & 62 & 45 \\
$G_5$        & 0  & 0   & 0  & 1  & 10 & 79 \\ \bottomrule
\end{tabular}
\end{table}

\begin{figure}[htbp]
    \includegraphics[width = \linewidth]{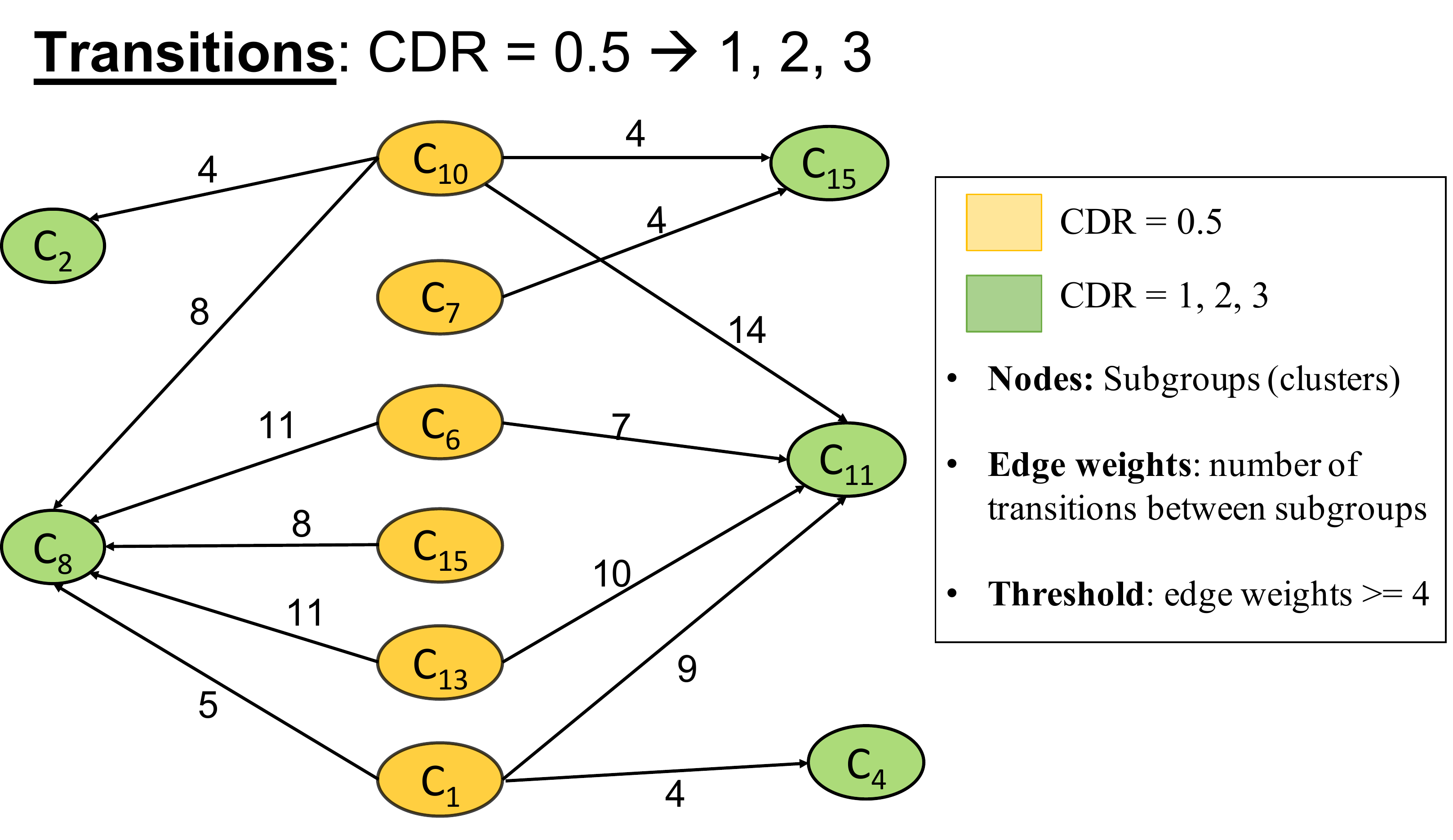}
    \caption{Graph showing transitions which are related to progression from CDR = 0.5 to higher CDR values 1, 2 and 3. The yellow and green nodes in the graph represent the subtypes with CDR = 0.5 and CDR = 1 visits respectively. For $C_{15}$, which has 2 CDR (0.5/1), we only considered the visits with CDR = 1. The edge weights represent the number of transitions from source to target. For example, edge weight of 4 between $C_{10} \rightarrow C_{15}$ means that there are 4 instances where patients were in $C_{10}$ in the current visit and in $C_{15}$ in their next visit. Only the transitions with edge weights >= 4 are shown here. }
    \label{fig:progression}
\end{figure}

Next, we examine how the dementia subtype group transitions shown above can be related to dementia progression. Figure \ref{fig:progression} shows the transitions from CDR = 0.5 to higher CDR score of 1, 2 and 3. Out of all the CDR 0.5 visits, individuals classified into dementia subtype $C_{10}$ and $C_13$ had the highest progression to CDR = 1 at their next visit. On the contrary, dementia subtypes like $C_7$ rarely progressed to higher CDR stages. This observation validates our hypothesis that individuals rated CDR = 0.5 are highly heterogeneous and have a variable rate of dementia progression.


\section{Discussion}

The goal of our study was to apply an unsupervised data-driven clustering approach to identify dementia subtypes within a specialty dementia clinic. We extracted outpatient office visits from an EHR, selected the 6 components of CDR as the optimal set of features after a feature selection step, and applied the K-means algorithm to identify 16 dementia subtypes. In contrast to prior work, our work focuses on both the interpretation of the identified subtypes based on their cognitive characteristics and analyzing how the subtypes play a role in the longitudinal progression trajectory of the disease.  

\subsection{Clustering on individual visits}

In this work, the unsupervised clustering analysis was performed with data from individual visits, rather than each patient, which assumed that each visit of the same patient is independent without any longitudinal relations between the consecutive visit. Our goal was to blind ourselves to the temporality (linkages) between visits, and later utilize the temporal aspect to investigate the relationship between the clusters (dementia subtypes). First, dementia often progresses over a decade, whereas our window of EHR data is a maximum of six years. Rigorous analysis of longitudinal data is only possible when information about all the visits of a particular dementia patient is available, and in this dataset, we are evaluating a cross-section of data from patients that encompasses different windows of their clinical care (e.g. some may have been seen in 2006-2014, others from 2016-2022). Since we had EHR data from 2012-2018 only, we decided to perform the clustering at the visit level. This approach also helps in increasing the volume of data available for clustering, since we had only 1845 patients in our dataset who satisfied the eligibility criteria. Since the units of each dementia subtype are individual visits and not patients, a patient can have the first visit in one dementia subtype and the next visit in the same or a different subtype. This approach not only allows us to analyze the relationship between dementia subtypes but also the differential progression trajectory of dementia patients. Although our experimental analyzes have been demonstrated in terms of dementia subtypes, our findings could be made more granular by focusing on the heterogeneity of individual patients. For example, when we say a particular dementia subtype with CDR = 0.5 has a higher chance of progression to CDR = 1, 2, 3 compared to the other subtypes, we can investigate the patients who have visits in that particular subtype and track their remaining visits across all subtypes. We believe that this visit-level approach is more robust and generalizable in terms of investigating heterogeneity and disease progression compared to the patient-level approach of clustering using only the baseline information of each dementia patient.

\subsection{Prioritizing early stage dementia subgroups for precision diagnostics}

In our analysis, we focused our attention on the visits with CDR = 0.5 to investigate the level of heterogeneity. From our observation in Figures 2-5, we can conclude very mild dementia (CDR = 0.5) is much more heterogeneous than more severe dementia. We defined the cognitive profile of each dementia subtype and examined both the inter-subtype and intra-subtype variability. For example, individuals in the same dementia subtype can have different scores in memory, orientation, judgement and problem solving and community affairs (intra-subtype variability). We also evaluated the variability amongst dementia subtypes in their performance on the individual component tasks for CDR. In Figure \ref{fig:cdr_components_cluster}, we can see that the dementia subtypes with more severe dementia (CDR = 1, 2, 3) have worse performance on CDR component tasks with lower variability. Improved classification of individuals with very mild dementia may enable more accurate diagnosis and prognosis, which may also be helpful in weighing the potential benefits of certain treatments.  For example, if a patient is very likely to progress to more severe dementia, a treatment with significant risks may have greater justification.

\subsection{Variability in CDR progression}

In Table \ref{tab:group_transition}, we observed two interesting patterns in terms of CDR progression: (i) regression in CDR global score, signifying that some patients improve and (ii) skipping CDR stages during progression, where a patient can skip a CDR stage during progression (e.g. move directly from CDR = 0.5 to 2 in their next visit). The regression in CDR score could be related to treatment of conditions that can cause or worsen dementia symptoms. For example, dementia specialists often discontinue medications that impair cognition, and patients often improve after these changes are made. Skipping CDR stages during progression could be related to actual rapid progression or a long interval between adjacent visits. Events like a stroke or a fall can significantly worsen the cognitive abilities of a patient, increasing the apparent rate of progression. Changes in the collateral source (caregiver, family member) who accompanies the patient in his/her follow-up visit and informs the clinician of the patient's condition can affect the CDR. This reflects a potential bias in the clinical rating of dementia.

\subsection{Limitations and scope for future work}

One of the limitations is the fact that we have tested our approach on a single EHR dataset. As part of future work, we plan to test the generalizability of our pipeline on additional datasets. Another limitation is the relatively small sample size of our dataset which restricts our ability to perform rigorous statistical analyzes on the variable rate of progression of dementia subtypes. The main motivation behind our analysis was to analyze the initial patterns of heterogeneity seen within a dementia cohort and possible solution of parsing the heterogeneity by characterizing the dementia subtypes. Applying this pipeline to a larger EHR dataset can allow us to estimate probabilistic estimates of disease progression, backed by rigorous statistical analysis. 

\section{Conclusion}

In this research, we applied an unsupervised data-driven clustering approach to EHR data from a dementia specialty clinic to identify dementia subtypes based on CDR component scores. Our goal was to characterize the cognitive profile of the dementia subtypes and analyze how the subtypes change over time. We observed that dementia subtypes that represented individuals with very mild dementia (CDR 0.5) had widely varying rates of transition to other subtypes.  Future work includes testing the generalizability of our proposed pipeline on additional datasets, and using a larger volume of EHR data to estimate probabilistic estimates of the variability between dementia subtypes both in terms of cognitive profile and disease progression.



\bibliographystyle{ACM-Reference-Format}
\bibliography{references}


\begin{thebibliography}{28}


\ifx \showCODEN    \undefined \def \showCODEN     #1{\unskip}     \fi
\ifx \showDOI      \undefined \def \showDOI       #1{#1}\fi
\ifx \showISBNx    \undefined \def \showISBNx     #1{\unskip}     \fi
\ifx \showISBNxiii \undefined \def \showISBNxiii  #1{\unskip}     \fi
\ifx \showISSN     \undefined \def \showISSN      #1{\unskip}     \fi
\ifx \showLCCN     \undefined \def \showLCCN      #1{\unskip}     \fi
\ifx \shownote     \undefined \def \shownote      #1{#1}          \fi
\ifx \showarticletitle \undefined \def \showarticletitle #1{#1}   \fi
\ifx \showURL      \undefined \def \showURL       {\relax}        \fi
\providecommand\bibfield[2]{#2}
\providecommand\bibinfo[2]{#2}
\providecommand\natexlab[1]{#1}
\providecommand\showeprint[2][]{arXiv:#2}

\bibitem[Davidson et~al\mbox{.}(2010)]%
        {davidson2010exploration}
\bibfield{author}{\bibinfo{person}{Julie~E Davidson},
  \bibinfo{person}{Michael~C Irizarry}, \bibinfo{person}{Bethany~C Bray},
  \bibinfo{person}{Sally Wetten}, \bibinfo{person}{Nicholas Galwey},
  \bibinfo{person}{Rachel Gibson}, \bibinfo{person}{Michael Borrie},
  \bibinfo{person}{Richard Delisle}, \bibinfo{person}{Howard~H Feldman},
  \bibinfo{person}{Ging-Yuek Hsiung}, {et~al\mbox{.}}}
  \bibinfo{year}{2010}\natexlab{}.
\newblock \showarticletitle{An exploration of cognitive subgroups in
  Alzheimer’s disease}.
\newblock \bibinfo{journal}{\emph{Journal of the International
  Neuropsychological Society}} \bibinfo{volume}{16}, \bibinfo{number}{2}
  (\bibinfo{year}{2010}), \bibinfo{pages}{233--243}.
\newblock


\bibitem[Dong et~al\mbox{.}(2015)]%
        {dong2015chimera}
\bibfield{author}{\bibinfo{person}{Aoyan Dong}, \bibinfo{person}{Nicolas
  Honnorat}, \bibinfo{person}{Bilwaj Gaonkar}, {and} \bibinfo{person}{Christos
  Davatzikos}.} \bibinfo{year}{2015}\natexlab{}.
\newblock \showarticletitle{CHIMERA: clustering of heterogeneous disease
  effects via distribution matching of imaging patterns}.
\newblock \bibinfo{journal}{\emph{IEEE transactions on medical imaging}}
  \bibinfo{volume}{35}, \bibinfo{number}{2} (\bibinfo{year}{2015}),
  \bibinfo{pages}{612--621}.
\newblock


\bibitem[Ferri et~al\mbox{.}(2005)]%
        {ferri2005global}
\bibfield{author}{\bibinfo{person}{Cleusa~P Ferri}, \bibinfo{person}{Martin
  Prince}, \bibinfo{person}{Carol Brayne}, \bibinfo{person}{Henry Brodaty},
  \bibinfo{person}{Laura Fratiglioni}, \bibinfo{person}{Mary Ganguli},
  \bibinfo{person}{Kathleen Hall}, \bibinfo{person}{Kazuo Hasegawa},
  \bibinfo{person}{Hugh Hendrie}, \bibinfo{person}{Yueqin Huang},
  {et~al\mbox{.}}} \bibinfo{year}{2005}\natexlab{}.
\newblock \showarticletitle{Global prevalence of dementia: a Delphi consensus
  study}.
\newblock \bibinfo{journal}{\emph{The lancet}} \bibinfo{volume}{366},
  \bibinfo{number}{9503} (\bibinfo{year}{2005}), \bibinfo{pages}{2112--2117}.
\newblock


\bibitem[F{\"o}rstl et~al\mbox{.}(1994)]%
        {forstl1994pathways}
\bibfield{author}{\bibinfo{person}{H F{\"o}rstl}, \bibinfo{person}{R Levy},
  \bibinfo{person}{A Burns}, \bibinfo{person}{P Luthert}, {and}
  \bibinfo{person}{N Cairns}.} \bibinfo{year}{1994}\natexlab{}.
\newblock \showarticletitle{Pathways and patterns of cell loss in verified
  Alzheimer's disease: a factor and cluster analysis of clinico-pathological
  subgroups}.
\newblock \bibinfo{journal}{\emph{Behavioural neurology}} \bibinfo{volume}{7},
  \bibinfo{number}{3-4} (\bibinfo{year}{1994}), \bibinfo{pages}{175--180}.
\newblock


\bibitem[Goyal et~al\mbox{.}(2018)]%
        {goyal2018characterizing}
\bibfield{author}{\bibinfo{person}{Devendra Goyal}, \bibinfo{person}{Donna
  Tjandra}, \bibinfo{person}{Raymond~Q Migrino}, \bibinfo{person}{Bruno
  Giordani}, \bibinfo{person}{Zeeshan Syed}, \bibinfo{person}{Jenna Wiens},
  \bibinfo{person}{Alzheimer's Disease~Neuroimaging Initiative},
  {et~al\mbox{.}}} \bibinfo{year}{2018}\natexlab{}.
\newblock \showarticletitle{Characterizing heterogeneity in the progression of
  Alzheimer's disease using longitudinal clinical and neuroimaging biomarkers}.
\newblock \bibinfo{journal}{\emph{Alzheimer's \& Dementia: Diagnosis,
  Assessment \& Disease Monitoring}}  \bibinfo{volume}{10}
  (\bibinfo{year}{2018}), \bibinfo{pages}{629--637}.
\newblock


\bibitem[Jack~Jr et~al\mbox{.}(2015)]%
        {jack2015different}
\bibfield{author}{\bibinfo{person}{Clifford~R Jack~Jr},
  \bibinfo{person}{Heather~J Wiste}, \bibinfo{person}{Stephen~D Weigand},
  \bibinfo{person}{David~S Knopman}, \bibinfo{person}{Michelle~M Mielke},
  \bibinfo{person}{Prashanthi Vemuri}, \bibinfo{person}{Val Lowe},
  \bibinfo{person}{Matthew~L Senjem}, \bibinfo{person}{Jeffrey~L Gunter},
  \bibinfo{person}{Denise Reyes}, {et~al\mbox{.}}}
  \bibinfo{year}{2015}\natexlab{}.
\newblock \showarticletitle{Different definitions of neurodegeneration produce
  similar amyloid/neurodegeneration biomarker group findings}.
\newblock \bibinfo{journal}{\emph{Brain}} \bibinfo{volume}{138},
  \bibinfo{number}{12} (\bibinfo{year}{2015}), \bibinfo{pages}{3747--3759}.
\newblock


\bibitem[Kumar et~al\mbox{.}(2021)]%
        {kumar2021machine}
\bibfield{author}{\bibinfo{person}{Sayantan Kumar}, \bibinfo{person}{Inez Oh},
  \bibinfo{person}{Suzanne Schindler}, \bibinfo{person}{Albert~M Lai},
  \bibinfo{person}{Philip~RO Payne}, {and} \bibinfo{person}{Aditi Gupta}.}
  \bibinfo{year}{2021}\natexlab{}.
\newblock \showarticletitle{Machine learning for modeling the progression of
  Alzheimer disease dementia using clinical data: a systematic literature
  review}.
\newblock \bibinfo{journal}{\emph{JAMIA open}} \bibinfo{volume}{4},
  \bibinfo{number}{3} (\bibinfo{year}{2021}), \bibinfo{pages}{ooab052}.
\newblock


\bibitem[Lam et~al\mbox{.}(2013)]%
        {lam2013clinical}
\bibfield{author}{\bibinfo{person}{Benjamin Lam}, \bibinfo{person}{Mario
  Masellis}, \bibinfo{person}{Morris Freedman}, \bibinfo{person}{Donald~T
  Stuss}, {and} \bibinfo{person}{Sandra~E Black}.}
  \bibinfo{year}{2013}\natexlab{}.
\newblock \showarticletitle{Clinical, imaging, and pathological heterogeneity
  of the Alzheimer's disease syndrome}.
\newblock \bibinfo{journal}{\emph{Alzheimer's research \& therapy}}
  \bibinfo{volume}{5}, \bibinfo{number}{1} (\bibinfo{year}{2013}),
  \bibinfo{pages}{1--14}.
\newblock


\bibitem[Landi et~al\mbox{.}(2020)]%
        {landi2020deep}
\bibfield{author}{\bibinfo{person}{Isotta Landi}, \bibinfo{person}{Benjamin~S
  Glicksberg}, \bibinfo{person}{Hao-Chih Lee}, \bibinfo{person}{Sarah Cherng},
  \bibinfo{person}{Giulia Landi}, \bibinfo{person}{Matteo Danieletto},
  \bibinfo{person}{Joel~T Dudley}, \bibinfo{person}{Cesare Furlanello}, {and}
  \bibinfo{person}{Riccardo Miotto}.} \bibinfo{year}{2020}\natexlab{}.
\newblock \showarticletitle{Deep representation learning of electronic health
  records to unlock patient stratification at scale}.
\newblock \bibinfo{journal}{\emph{NPJ digital medicine}} \bibinfo{volume}{3},
  \bibinfo{number}{1} (\bibinfo{year}{2020}), \bibinfo{pages}{1--11}.
\newblock


\bibitem[Likas et~al\mbox{.}(2003)]%
        {likas2003global}
\bibfield{author}{\bibinfo{person}{Aristidis Likas}, \bibinfo{person}{Nikos
  Vlassis}, {and} \bibinfo{person}{Jakob~J Verbeek}.}
  \bibinfo{year}{2003}\natexlab{}.
\newblock \showarticletitle{The global k-means clustering algorithm}.
\newblock \bibinfo{journal}{\emph{Pattern recognition}} \bibinfo{volume}{36},
  \bibinfo{number}{2} (\bibinfo{year}{2003}), \bibinfo{pages}{451--461}.
\newblock


\bibitem[Malpas(2016)]%
        {malpas2016structural}
\bibfield{author}{\bibinfo{person}{Charles~B Malpas}.}
  \bibinfo{year}{2016}\natexlab{}.
\newblock \showarticletitle{Structural neuroimaging correlates of cognitive
  status in older adults: a person-oriented approach}.
\newblock \bibinfo{journal}{\emph{Journal of Clinical Neuroscience}}
  \bibinfo{volume}{30} (\bibinfo{year}{2016}), \bibinfo{pages}{77--82}.
\newblock


\bibitem[Morris(1991)]%
        {morris1991clinical}
\bibfield{author}{\bibinfo{person}{John~C Morris}.}
  \bibinfo{year}{1991}\natexlab{}.
\newblock \showarticletitle{The clinical dementia rating (cdr): Current version
  and}.
\newblock \bibinfo{journal}{\emph{Young}}  \bibinfo{volume}{41}
  (\bibinfo{year}{1991}), \bibinfo{pages}{1588--1592}.
\newblock


\bibitem[Murray et~al\mbox{.}(2011)]%
        {murray2011neuropathologically}
\bibfield{author}{\bibinfo{person}{Melissa~E Murray}, \bibinfo{person}{Neill~R
  Graff-Radford}, \bibinfo{person}{Owen~A Ross}, \bibinfo{person}{Ronald~C
  Petersen}, \bibinfo{person}{Ranjan Duara}, {and} \bibinfo{person}{Dennis~W
  Dickson}.} \bibinfo{year}{2011}\natexlab{}.
\newblock \showarticletitle{Neuropathologically defined subtypes of Alzheimer's
  disease with distinct clinical characteristics: a retrospective study}.
\newblock \bibinfo{journal}{\emph{The Lancet Neurology}} \bibinfo{volume}{10},
  \bibinfo{number}{9} (\bibinfo{year}{2011}), \bibinfo{pages}{785--796}.
\newblock


\bibitem[Nettiksimmons et~al\mbox{.}(2014)]%
        {nettiksimmons2014biological}
\bibfield{author}{\bibinfo{person}{Jasmine Nettiksimmons},
  \bibinfo{person}{Charles DeCarli}, \bibinfo{person}{Susan Landau},
  \bibinfo{person}{Laurel Beckett}, \bibinfo{person}{Alzheimer's
  Disease~Neuroimaging Initiative}, {et~al\mbox{.}}}
  \bibinfo{year}{2014}\natexlab{}.
\newblock \showarticletitle{Biological heterogeneity in ADNI amnestic mild
  cognitive impairment}.
\newblock \bibinfo{journal}{\emph{Alzheimer's \& Dementia}}
  \bibinfo{volume}{10}, \bibinfo{number}{5} (\bibinfo{year}{2014}),
  \bibinfo{pages}{511--521}.
\newblock


\bibitem[Noh et~al\mbox{.}(2014)]%
        {noh2014anatomical}
\bibfield{author}{\bibinfo{person}{Young Noh}, \bibinfo{person}{Seun Jeon},
  \bibinfo{person}{Jong~Min Lee}, \bibinfo{person}{Sang~Won Seo},
  \bibinfo{person}{Geon~Ha Kim}, \bibinfo{person}{Hanna Cho},
  \bibinfo{person}{Byoung~Seok Ye}, \bibinfo{person}{Cindy~W Yoon},
  \bibinfo{person}{Hee~Jin Kim}, \bibinfo{person}{Juhee Chin}, {et~al\mbox{.}}}
  \bibinfo{year}{2014}\natexlab{}.
\newblock \showarticletitle{Anatomical heterogeneity of Alzheimer disease:
  based on cortical thickness on MRIs}.
\newblock \bibinfo{journal}{\emph{Neurology}} \bibinfo{volume}{83},
  \bibinfo{number}{21} (\bibinfo{year}{2014}), \bibinfo{pages}{1936--1944}.
\newblock


\bibitem[Poulakis et~al\mbox{.}(2018)]%
        {poulakis2018heterogeneous}
\bibfield{author}{\bibinfo{person}{Konstantinos Poulakis},
  \bibinfo{person}{Joana~B Pereira}, \bibinfo{person}{Patrizia Mecocci},
  \bibinfo{person}{Bruno Vellas}, \bibinfo{person}{Magda Tsolaki},
  \bibinfo{person}{Iwona K{\l}oszewska}, \bibinfo{person}{Hilkka Soininen},
  \bibinfo{person}{Simon Lovestone}, \bibinfo{person}{Andrew Simmons},
  \bibinfo{person}{Lars-Olof Wahlund}, {et~al\mbox{.}}}
  \bibinfo{year}{2018}\natexlab{}.
\newblock \showarticletitle{Heterogeneous patterns of brain atrophy in
  Alzheimer's disease}.
\newblock \bibinfo{journal}{\emph{Neurobiology of aging}}  \bibinfo{volume}{65}
  (\bibinfo{year}{2018}), \bibinfo{pages}{98--108}.
\newblock


\bibitem[Price et~al\mbox{.}(2015)]%
        {price2015dissociating}
\bibfield{author}{\bibinfo{person}{Catherine~C Price}, \bibinfo{person}{Jared~J
  Tanner}, \bibinfo{person}{Ilona~M Schmalfuss}, \bibinfo{person}{Babette
  Brumback}, \bibinfo{person}{Kenneth~M Heilman}, {and}
  \bibinfo{person}{David~J Libon}.} \bibinfo{year}{2015}\natexlab{}.
\newblock \showarticletitle{Dissociating statistically-determined Alzheimer’s
  disease/vascular dementia neuropsychological syndromes using white and gray
  neuroradiological parameters}.
\newblock \bibinfo{journal}{\emph{Journal of Alzheimer's Disease}}
  \bibinfo{volume}{48}, \bibinfo{number}{3} (\bibinfo{year}{2015}),
  \bibinfo{pages}{833--847}.
\newblock


\bibitem[Ryan et~al\mbox{.}(2018)]%
        {ryan2018phenotypic}
\bibfield{author}{\bibinfo{person}{Joanne Ryan}, \bibinfo{person}{Peter
  Fransquet}, \bibinfo{person}{Jo Wrigglesworth}, {and} \bibinfo{person}{Paul
  Lacaze}.} \bibinfo{year}{2018}\natexlab{}.
\newblock \showarticletitle{Phenotypic heterogeneity in dementia: a challenge
  for epidemiology and biomarker studies}.
\newblock \bibinfo{journal}{\emph{Frontiers in public health}}
  \bibinfo{volume}{6} (\bibinfo{year}{2018}), \bibinfo{pages}{181}.
\newblock


\bibitem[Scheltens et~al\mbox{.}(2016)]%
        {scheltens2016identification}
\bibfield{author}{\bibinfo{person}{Nienke~ME Scheltens},
  \bibinfo{person}{Francisca Galindo-Garre}, \bibinfo{person}{Yolande~AL
  Pijnenburg}, \bibinfo{person}{Annelies~E van~der Vlies},
  \bibinfo{person}{Lieke~L Smits}, \bibinfo{person}{Teddy Koene},
  \bibinfo{person}{Charlotte~E Teunissen}, \bibinfo{person}{Frederik Barkhof},
  \bibinfo{person}{Mike~P Wattjes}, \bibinfo{person}{Philip Scheltens},
  {et~al\mbox{.}}} \bibinfo{year}{2016}\natexlab{}.
\newblock \showarticletitle{The identification of cognitive subtypes in
  Alzheimer's disease dementia using latent class analysis}.
\newblock \bibinfo{journal}{\emph{Journal of Neurology, Neurosurgery \&
  Psychiatry}} \bibinfo{volume}{87}, \bibinfo{number}{3}
  (\bibinfo{year}{2016}), \bibinfo{pages}{235--243}.
\newblock


\bibitem[Scheltens et~al\mbox{.}(2017)]%
        {scheltens2017cognitive}
\bibfield{author}{\bibinfo{person}{Nienke~ME Scheltens},
  \bibinfo{person}{Betty~M Tijms}, \bibinfo{person}{Teddy Koene},
  \bibinfo{person}{Frederik Barkhof}, \bibinfo{person}{Charlotte~E Teunissen},
  \bibinfo{person}{Steffen Wolfsgruber}, \bibinfo{person}{Michael Wagner},
  \bibinfo{person}{Johannes Kornhuber}, \bibinfo{person}{Oliver Peters},
  \bibinfo{person}{Brendan~I Cohn-Sheehy}, {et~al\mbox{.}}}
  \bibinfo{year}{2017}\natexlab{}.
\newblock \showarticletitle{Cognitive subtypes of probable Alzheimer's disease
  robustly identified in four cohorts}.
\newblock \bibinfo{journal}{\emph{Alzheimer's \& Dementia}}
  \bibinfo{volume}{13}, \bibinfo{number}{11} (\bibinfo{year}{2017}),
  \bibinfo{pages}{1226--1236}.
\newblock


\bibitem[Tibshirani et~al\mbox{.}(2001)]%
        {tibshirani2001estimating}
\bibfield{author}{\bibinfo{person}{Robert Tibshirani},
  \bibinfo{person}{Guenther Walther}, {and} \bibinfo{person}{Trevor Hastie}.}
  \bibinfo{year}{2001}\natexlab{}.
\newblock \showarticletitle{Estimating the number of clusters in a data set via
  the gap statistic}.
\newblock \bibinfo{journal}{\emph{Journal of the Royal Statistical Society:
  Series B (Statistical Methodology)}} \bibinfo{volume}{63},
  \bibinfo{number}{2} (\bibinfo{year}{2001}), \bibinfo{pages}{411--423}.
\newblock


\bibitem[Van~der Maaten and Hinton(2008)]%
        {van2008visualizing}
\bibfield{author}{\bibinfo{person}{Laurens Van~der Maaten} {and}
  \bibinfo{person}{Geoffrey Hinton}.} \bibinfo{year}{2008}\natexlab{}.
\newblock \showarticletitle{Visualizing data using t-SNE.}
\newblock \bibinfo{journal}{\emph{Journal of machine learning research}}
  \bibinfo{volume}{9}, \bibinfo{number}{11} (\bibinfo{year}{2008}).
\newblock


\bibitem[Varol et~al\mbox{.}(2017)]%
        {varol2017hydra}
\bibfield{author}{\bibinfo{person}{Erdem Varol}, \bibinfo{person}{Aristeidis
  Sotiras}, \bibinfo{person}{Christos Davatzikos}, \bibinfo{person}{Alzheimer's
  Disease~Neuroimaging Initiative}, {et~al\mbox{.}}}
  \bibinfo{year}{2017}\natexlab{}.
\newblock \showarticletitle{HYDRA: Revealing heterogeneity of imaging and
  genetic patterns through a multiple max-margin discriminative analysis
  framework}.
\newblock \bibinfo{journal}{\emph{Neuroimage}}  \bibinfo{volume}{145}
  (\bibinfo{year}{2017}), \bibinfo{pages}{346--364}.
\newblock


\bibitem[Vogt and Nagel(1992)]%
        {vogt1992cluster}
\bibfield{author}{\bibinfo{person}{Wolfgang Vogt} {and}
  \bibinfo{person}{Dorothea Nagel}.} \bibinfo{year}{1992}\natexlab{}.
\newblock \showarticletitle{Cluster analysis in diagnosis}.
\newblock \bibinfo{journal}{\emph{Clinical Chemistry}} \bibinfo{volume}{38},
  \bibinfo{number}{2} (\bibinfo{year}{1992}), \bibinfo{pages}{182--198}.
\newblock


\bibitem[Wallin et~al\mbox{.}(2011)]%
        {wallin2011galantamine}
\bibfield{author}{\bibinfo{person}{{\AA}sa~K Wallin}, \bibinfo{person}{Carina
  Wattmo}, {and} \bibinfo{person}{Lennart Minthon}.}
  \bibinfo{year}{2011}\natexlab{}.
\newblock \showarticletitle{Galantamine treatment in Alzheimer’s disease:
  response and long-term outcome in a routine clinical setting}.
\newblock \bibinfo{journal}{\emph{Neuropsychiatric disease and treatment}}
  \bibinfo{volume}{7} (\bibinfo{year}{2011}), \bibinfo{pages}{565}.
\newblock


\bibitem[Xu et~al\mbox{.}(2020)]%
        {xu2020data}
\bibfield{author}{\bibinfo{person}{Jie Xu}, \bibinfo{person}{Fei Wang},
  \bibinfo{person}{Zhenxing Xu}, \bibinfo{person}{Prakash Adekkanattu},
  \bibinfo{person}{Pascal Brandt}, \bibinfo{person}{Guoqian Jiang},
  \bibinfo{person}{Richard~C Kiefer}, \bibinfo{person}{Yuan Luo},
  \bibinfo{person}{Chengsheng Mao}, \bibinfo{person}{Jennifer~A Pacheco},
  {et~al\mbox{.}}} \bibinfo{year}{2020}\natexlab{}.
\newblock \showarticletitle{Data-driven discovery of probable Alzheimer's
  disease and related dementia subphenotypes using electronic health records}.
\newblock \bibinfo{journal}{\emph{Learning Health Systems}}
  \bibinfo{volume}{4}, \bibinfo{number}{4} (\bibinfo{year}{2020}),
  \bibinfo{pages}{e10246}.
\newblock


\bibitem[Zanetti et~al\mbox{.}(2006)]%
        {zanetti2006mild}
\bibfield{author}{\bibinfo{person}{Mariella Zanetti}, \bibinfo{person}{Claudia
  Ballabio}, \bibinfo{person}{Carlo Abbate}, \bibinfo{person}{Chiara Cutaia},
  \bibinfo{person}{Carlo Vergani}, {and} \bibinfo{person}{Luigi
  Bergamaschini}.} \bibinfo{year}{2006}\natexlab{}.
\newblock \showarticletitle{Mild cognitive impairment subtypes and vascular
  dementia in community-dwelling elderly people: a 3-year follow-up study}.
\newblock \bibinfo{journal}{\emph{Journal of the American Geriatrics Society}}
  \bibinfo{volume}{54}, \bibinfo{number}{4} (\bibinfo{year}{2006}),
  \bibinfo{pages}{580--586}.
\newblock


\bibitem[Zhou and Saghapour(2021)]%
        {zhou2021imputehr}
\bibfield{author}{\bibinfo{person}{Yi-Hui Zhou} {and} \bibinfo{person}{Ehsan
  Saghapour}.} \bibinfo{year}{2021}\natexlab{}.
\newblock \showarticletitle{ImputEHR: a visualization tool of imputation for
  the prediction of biomedical data}.
\newblock \bibinfo{journal}{\emph{Frontiers in Genetics}}  \bibinfo{volume}{12}
  (\bibinfo{year}{2021}).
\newblock


\end{thebibliography}


\end{document}